\documentclass[runningheads]{llncs}
\usepackage{amssymb}
\usepackage{graphicx}
\usepackage{amsmath}
\usepackage{multirow}
\usepackage{multicol}
\usepackage{makecell}
\usepackage{amssymb}
\usepackage{siunitx}
\usepackage{tabularx}
\usepackage[misc]{ifsym}
\usepackage{subfig}  
\usepackage{gensymb}

\begin{document}
%
% 隐藏基金信息
% \title{MSF-Net: Multi-Stage Feature Extraction and Fusion for Robust Photometric Stereo\thanks{
% Supported by Key R\&D Program of Shandong Province, China (Grant No.anonymous). }}
% 
% 
% 隐藏基金信息
\title{MSF-Net: Multi-Stage Feature Extraction and Fusion for Robust Photometric Stereo\thanks{
Supported by Key R\&D Program of Shandong Province, China (Grant No.
2024ZLGX06). }}
\titlerunning{MSF-Net for Robust PS}

\author{
Shiyu Qin\inst{1} \and 
Zhihao Cai\inst{1} \and 
Kaixuan Wang\inst{1} \and 
Lin Qi\inst{1}{(\textrm{\Letter})} \and % 第四作者（通讯作者）
Junyu Dong\inst{1} % 第五作者
}
\authorrunning{Qin et al.}
\institute{
School of Computer Science and Technology, 
Ocean University of China, 
Qingdao 266000, China \\
\email{qsy8264@stu.ouc.edu.cn}, 
\email{caizhihao@stu.ouc.edu.cn}, 
\email{wangkaixuan@stu.ouc.edu.cn}, 
\email{qilin@ouc.edu.cn}, % 通讯邮箱标注
\email{dongjunyu@ouc.edu.cn}
}

\maketitle              

\begin{abstract}

% Photometric stereo is a technique aimed at determining surface normals through the utilization of shading cues derived from images taken under different lighting conditions.
% However, existing learning-based approaches often fail to accurately capture features at multiple stages and do not adequately promote interaction between these features.
% Consequently, these models tend to extract redundant features, especially in areas with intricate details such as wrinkles and edges.
% % 
% To tackle these issues, we propose MSF-Net, a novel framework for extracting information at multiple stages, paired with selective update strategy, aiming to extract high-quality feature information, which is critical for accurate normal construction.
% Additionally, we have developed a feature fusion module to improve the interplay among different features.
% % 
% % Abstract里对数据集不需要添加引用吗？？ 摘要里就不引用了。
% % 师姐论文也没提其它数据集，那我也不提了
% Experimental results on the DiLiGenT benchmark show that our proposed MSF-Net significantly surpasses previous state-of-the-art methods in the accuracy of surface normal estimation.
Photometric stereo is a technique aimed at determining surface normals through the utilization of shading cues derived from images taken under different lighting conditions. However, existing learning-based approaches often fail to accurately capture features at multiple stages and do not adequately promote interaction between these features. Consequently, these models tend to extract redundant features, especially in areas with intricate details such as wrinkles and edges. To tackle these issues, we propose MSF-Net, a novel framework for extracting information at multiple stages, paired with selective update strategy, aiming to extract high-quality feature information, which is critical for accurate normal construction. Additionally, we have developed a feature fusion module to improve the interplay among different features. Experimental results on the DiLiGenT benchmark show that our proposed MSF-Net significantly surpasses previous state-of-the-art methods in the accuracy of surface normal estimation.
\keywords{photometric stereo  \and multi-stage feature \and selective update.}
\end{abstract}
\section{Introduction}
Photometric stereo (PS), introduced by Woodham \cite{Woodham_1980}, is a technique for recovering surface normals of an object from multiple images captured under varying illumination conditions from a fixed viewpoint. 
Due to its ability to reconstruct fine surface details, PS has found widespread applications in many fields \cite{application_1_Fan_Qi_Ju_Dong_Yu_2017} \cite{application_2-Ren_Wang_Xiao_Chen_Fu_2019}.
% 
% PS的问题
However, photometric stereo is inherently ill-posed due to unknown surface reflectance and global illumination effects. 
Traditional approaches typically assume Lambertian reflectance to simplify computation of surface normals using least-squares methods \cite{Woodham_1980}.
However, real-world objects often exhibit non-Lambertian properties like specular highlights and interreflections, invalidating this assumption. 
While methods have been proposed to address these effects—including outlier rejection \cite{reject_1-Mukaigawa_Ishii_Shakunaga_2007} \cite{reject_2-4760144} and approximate Bidirectional Reflectance Distribution Functions (BRDFs) \cite{BRDF_1-fShi_Tan_Matsushita_Ikeuchi_2014} \cite{BRDF_2-Holroyd_Lawrence_Humphreys_Zickler_2008}—they lack generality across diverse materials and lighting conditions.

% 
% ---基于深度学习的PS方法

Recent progress in deep learning has significantly augmented the field of computer vision, leading to performance enhancements across a range of tasks. 
Many deep learning models have now successfully improved accuracy in estimating surface normals.
% 
% ---讲讲特征提取能力，引出多次法线回归
% DPSN
% 不能利用表面点周围的信息-->提取信息能力不好
The early work learned the mapping between reflection observations and surface normals using a fully connected network \cite{DPSN-Santo_Samejima_Sugano_Shi_Matsushita_2017}, which requires a fixed order and a predefined number of input images during both training and testing phases, thus constraining its adaptability and generalization capability.
% 
% PS-FCN
% Guanyin Chen et al.
While the milestone PS-FCN framework \cite{PSFCN_Chen_2018} introduces a flexible fully convolutional architecture capable of processing arbitrary unordered inputs, its reliance on max-pooling for feature fusion inherently compromises the retention of informative non-maximal features crucial for precise surface normal reconstruction \cite{PSMF_PSN-10301617}.
% 
% 
% ---现有工作的不足 
While existing models maximize information extraction through dense convolutions, this inadvertently introduces feature redundancy that compromises representation robustness and impairs surface normal estimation accuracy. 
Our suggested selective update strategy incrementally optimizes feature extractors, progressively enhancing feature precision and quality.
 
% 
% ---讲讲不同特征图的交互，引出特征融合模块
% 
% NormAttention-PSN
%

In order to improve the ability of models to deal with complex regions, 
Ju et al. proposed NormAttention-PSN, which employs innovative normalization techniques to minimize the impact of specular highlights and cast shadows \cite{NA_PSN-Ju_Shi_Jian_Qi_Dong_Lam_Kenneth}.
% 要说出问题（自己想的问题。。）
However, the computational operations involved in this model are notably complex.
% CNN-PS
% 只用了图像间信息
Ikehata et al. proposed CNN-PS, which maps information from input images into an observation map and significantly improves the accuracy of predicted surface normals \cite{CNN_PS_Ikehata_2018}.
% 问题
However, since the prediction is based on inter-image information, it is difficult to maintain good performance when the number of input images varies greatly \cite{PSMF_PSN-10301617}.
% 
% IGA-PSN
% 既褒又贬的亚子，主要提出本文模型还不错，大道至简
IGA-PSN \cite{IGA-PSN_10.1007/978-981-96-0122-6_25} integrates gradient information, amplifying the model's ability to utilize high-frequency image details.
Although effective, this comes with the trade-off of increased complexity and number of parameters, impacting computational efficiency. 
% 
% ---现有工作的不足 
% 重点是说其它模型对特征的利用、交互等不够充分，引出我的。
Deep convolutional architectures extract complex features at the cost of progressively discarding crucial original characteristics in deeper layers.
% 提一句我的方案
We propose a feature fusion module to counter this degradation by establishing inter-image feature interactions, enhancing surface normal reconstruction fidelity.
% Our code and models can be found at https://github.com/RainLikeCodes/MSF-Net.

\section{Related Work}
Considering a pixel located on a non-Lambertian surface illuminated by a directional light $l$ with a surface normal vector $n$.
When a linear-response camera captures an image of this surface in the view direction $v$, the pixel-measured intensity $i$ in image $I$ can be approximated as follows:

\begin{equation}
i = \rho(\mathbf{n},\mathbf{l},\mathbf{v}) \cdot \max \left\{\mathbf{n}^{\rm{T}}\mathbf{l},0 \right\}+ \varepsilon
\label{eq:imaging_model}
\end{equation}
where $\rho$ represents the BRDF, and $\max\left\{\mathbf{n}^{\rm{T}}\mathbf{l}, 0\right\}$ denotes the attached shadows, and $\varepsilon$ represents noise sources, such as cast shadows and inter-reflections \cite{inter_reflection-Nayar_Ikeuchi_Kanade_1991}. 
Under the simplifying assumption of Lambertian reflectance, the general BRDF $\rho$ is considerd as an constant, the error item is ignored, and the surface normals of general objects can be computed using no less than three images \cite{Woodham_1980}.
Unfortunately, most real-world objects do not exhibit Lambertian reflection, and the general BRDF $\rho$ and the noise $\varepsilon$ are unknown.
% 

% 简化版
In learning-based photometric stereo, \cite{DPSN-Santo_Samejima_Sugano_Shi_Matsushita_2017} pioneered deep neural networks for surface normal prediction but required ordered lighting directions. Subsequent research circumvented this limitation through full-pixel and per-pixel paradigms. 
Full-pixel methods \cite{Woodham_1980,mf_psn-Liu_Ju_Jian_Gao_Rao_Hu_Dong_2022,NA_PSN-Ju_Shi_Jian_Qi_Dong_Lam_Kenneth} aggregate features via max-pooling to learn pixel intensity variations. 
Per-pixel approaches \cite{CNN_PS_Ikehata_2018,PX_Net-Logothetis_Budvytis_Mecca_Cipolla_2020,ps_transformer-unknown} leverage observation maps to encode intensity variations across lighting conditions per pixel. 
GPS-Net \cite{GPS_Net-Yao_Li_Fu_Hu_Shi_2020} emerges as a hybrid solution integrating both paradigms.

Despite these innovations, contemporary methods suffer from feature redundancy due to stacked convolution operations, while overlooking strategic feature extraction design.

\section{Proposed Method}

\subsection{Image Preprocessing}
The reflection of light on opaque and non-flat surfaces complicates the task of distinguishing whether pixel value variations are due to structural or textural changes. 
Data normalization strategy was therefore proposed \cite{psfcn-n_Chen2020DeepPS}:

\begin{equation}
    i^{'}_{j}=\frac{i_{j}}{\sqrt{i^{2}_1+i^{2}_2+\ldots+i^{2}_n} }
    \label{norm_non_lam}
\end{equation}

The $i_j$ in the numerator represents the pixel value in the j-th image,
the denominator is the square root of the sum of squared pixel values at each location.
$n$ is the number of images, and $j \in \{1, 2, \ldots, n\}$. 
The $i^{'}_j$ represents the normalized pixel value of the j-th image. 
The normalization operation is performed independently for the RGB channels of the image.
% 
% 简化版
Assuming Lambertian reflection, Eq. \ref{eq:imaging_model} simplifies to $i_j = \rho \max(\mathbf{n}^{\rm{T}}\mathbf{l},0)$. Substituting Eq.~\ref{norm_non_lam} helps mitigate spatial BRDF variability from $\rho$. 
For non-Lambertian surfaces under directional lighting, low-frequency regions resemble Lambertian surfaces, while feature fusion disregards specular highlights and shadows, retaining salient features \cite{psfcn-n_Chen2020DeepPS}. Thus, normalization can be applied to non-Lambertian surfaces to handle spatially varying BRDFs, especially in regions with complex structures or textures.

% 模型结构图
\begin{figure}[t]
    \centering
    \includegraphics[width=\textwidth]{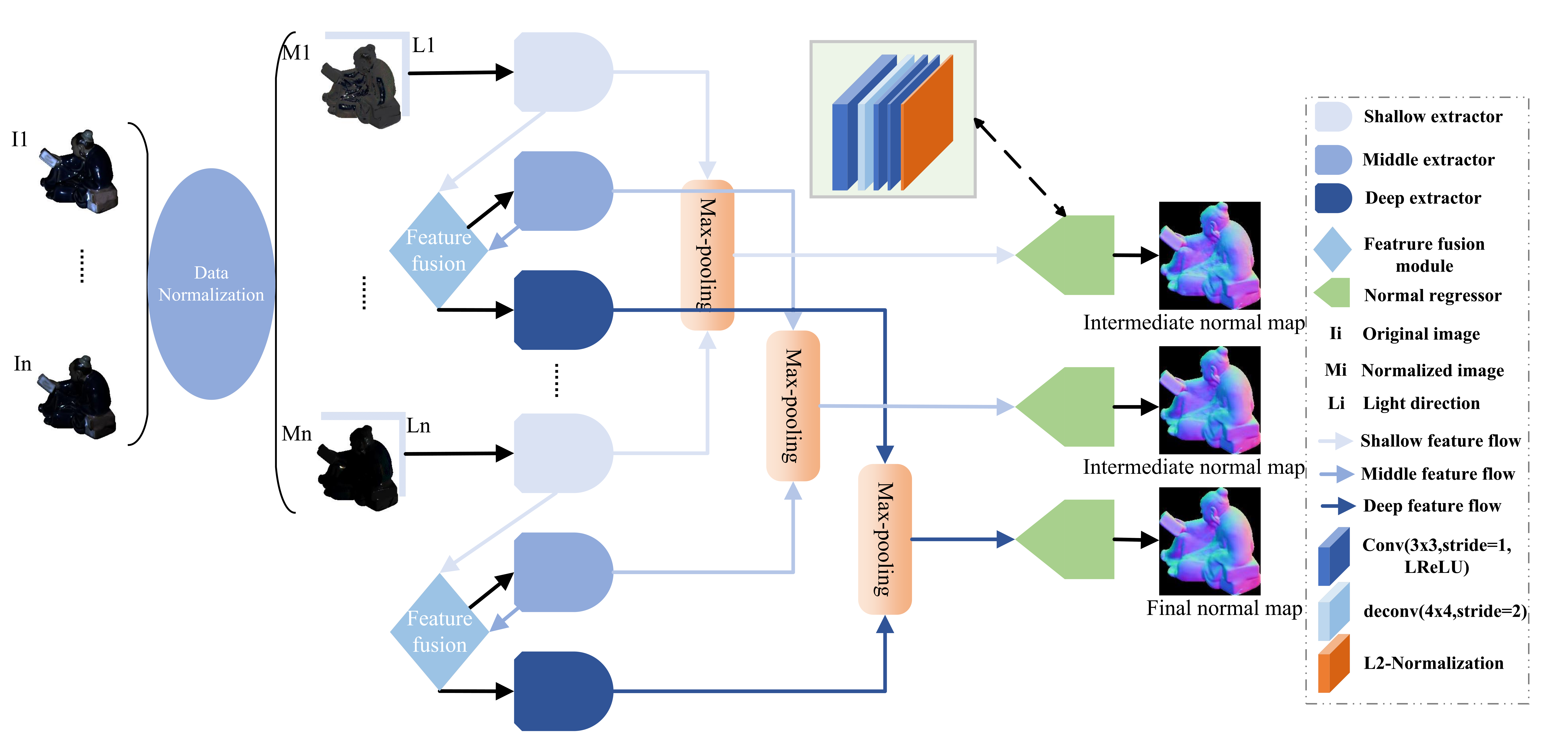}
    \caption{
    Network architecture of the proposed MSF-Net. 
The final normal map is determined solely by the output of the deep extractor.
    }
    \label{model_architecture}
\end{figure}

\subsection{MSF-Net}
The model we designed, shown in Fig. \ref{model_architecture}, includes three key components:
a shared-weight multi-stage feature extractor,
a shared-weight feature fusion module,
and a normal regression module.
In our proposed model, the shared-weight multi-stage feature extractor, hereafter referred to as the feature extractor, consists of three sub-networks: the shallow extractor, the middle extractor, and the deep extractor. These sub-networks extract feature information at different stages of the processing pipeline.

\subsubsection{Multi-stage Feature Extraction}
\begin{figure}
    \centering
    \includegraphics[width=0.8\textwidth]{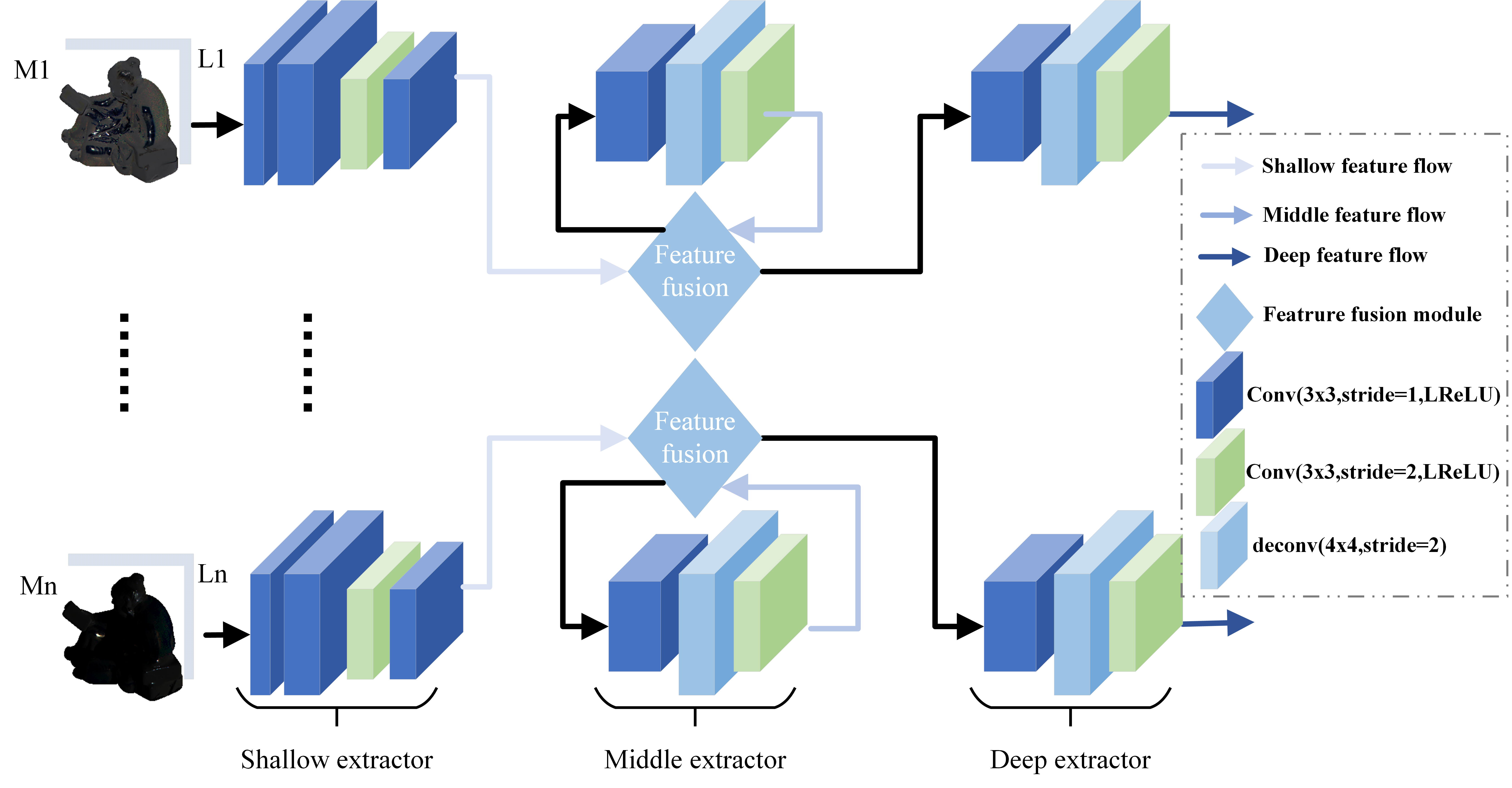}
    \caption{Architecture of the shared-weight multi-stage feature extractor. }
    \label{extractor_architecture}
\end{figure}

Our method decomposes feature extraction into three dedicated stages (shallow, middle, deep), each with specialized feature extractors. 
Fig. \ref{extractor_architecture} illustrates the architecture of the shared-wight multi-stage feature extractor we designed, which reveals the entire feature extraction process.

% 捋一下特征流动过程。
In the shallow stage, the feature extractor processes a set of image-light pairs to generate corresponding feature maps.
Since shallow layers perform relatively simple processing, these maps retain the input’s original characteristics without extensive mixing or aggregation typical in deeper layers.
Although limited feature diversity may result from a single extraction module, these features are essential for both initial normal estimation and subsequent processing.

Each predicted normal map is compared to ground truth (GT), optimizing the extraction module via the resulting loss.
The feature processing procedure of the middle stage is the same as that of the shallow stage, but input is feature maps output by shallow stage.
The process is represented by the following equations:
% 
% 插入公式
\begin{equation}
\begin{aligned}
% \label{eq:}
f^{M} &= Maxpool(f_{1},f_{2},\cdot,f_{n}) \\
N^{t} &= Reg(f^{M}) \\
f^{n} &= Fusion(f_{1},f_{2},\cdot,f_{n}) \\
% F^{'}_{i} &= Concat(f_{i},F_{i})
\end{aligned}
\end{equation}
where $Maxpool(\cdot)$ denotes max-pooling, $f_{i}$ is the $i$-th input at that stage, 
$f^{M}$ is the max-pooling result,
$Reg(\cdot)$ denotes the regression module,
$N^{t}$ is the predicted normal map, 
$Fusion(\cdot)$ indicates our feature fusion module, 
and $f^{n}$ is the output of the feature fusion module feeding the next stage.

Using all prior-stage feature maps – rather than only the max-pooled result – captures comprehensive information. 
This preserves critical non-maximal features for normal reconstruction, avoiding relevant detail loss.
After multi-stage extraction, the deep extractor's final feature maps contain rich high-level features. 
Their max-pooling captures prominent features for high-quality normal map reconstruction.

% 特征提取器结构图

\subsubsection{Multi-stage Feature Fusion}
Surface shading variations encoded in multi-light images provide fundamental cues for normal estimation \cite{PSMF_PSN-10301617}. However, effectively leveraging these cues demands explicit modeling of relationships across lighting conditions, a challenge inherent to photometric stereo. Conventional Convolutional Neural Networks (CNNs) extract features effectively within individual images but fail to fully exploit the relationships between features across the image set, causing critical shadow and highlight interactions to be overlooked.

To bridge this gap, our feature fusion module quantifies the similarity between two input feature maps using a cosine similarity measure, subsequently determining the extent to which each feature map's unique information should be retained based on the calculated similarity. 
The process can be formulated as:

\begin{equation}
\begin{aligned}
f^{M} &= Maxpool(f_{1},f_{2},\cdots,f_{n}) \\
f^{'}_i &= BN(Conv_1(GELU(f_i))) \\
f^{M'} &= BN(Conv_1(GELU(f^M))) \\
sim &= \sigma\left(\sum_{i} f^{'}_i \cdot f^{M'}\right) \\
% F_i &= (1 - sim) \cdot f_i + sim \cdot f^M \\
Out &= Concat((1 - sim) \cdot f_i + sim \cdot f^M, f_i)
\end{aligned}
\end{equation}
where $f_{i}$ denotes the i-th feature map at this stage, 
$f^{M}$ is the max-pooling result of all these feature maps.  
$GELU(\cdot)$ represents the GELU function,
$Conv_{1}(\cdot)$ denotes a 1 × 1 convolution, 
$BN(\cdot)$ stands for Batch Normalization, 
and $\sigma$ is the Sigmoid function,
Concat($\cdot$) represents the concatenate operation.
$Out$ is the output of the feature fusion module.

This adaptive fusion achieves three critical advances: first, it maintains sensitivity to subtle illumination-specific details that indicate surface orientation; second, it suppresses redundant features through cross-image consensus; third, and most significantly, it constructs an explicit reasoning pathway between lighting conditions. By dynamically reconciling diverse shading evidence—from deep shadows to specular highlights—the module transforms isolated per-image observations into a coherent physical understanding of surface geometry.

\subsubsection{Selective Update Strategy}
Our model employs the standard cosine similarity loss to supervise training, which directly measures the angular alignment between predicted and ground truth normals:
\begin{equation}
    Loss = \frac{1}{HW}\sum_{i,j}^{HW} (1-cos( \tilde{N}_{i,j} \cdot N_{i,j}))
\end{equation}
where $\tilde{N}_{i, j}$ and $N_{i, j}$ denote the predicted normal and the ground truth normal, respectively, at pixel (i, j).
We compute $Loss$ using output of extractor and GT at different stages, which drives our selective update strategy.

In the shallow stage, $Loss$ guides refinement of primary feature extraction. 
Shallow layers extract illumination-invariant geometry signatures, such as persistent edges under varying light directions, which serve as robust anchors against photometric noise.
% Since accurate shallow features form the essential foundation (e.g., edge information) for subsequent processing, this optimization focuses exclusively on the shallow extractor to ensure clean information propagation.

During middle-stage updates, we compute $Loss$ to adjust module weights but freeze shallow extractor gradients. 
This constraint prevents backpropagation from corrupting carefully distilled shallow extractor's capacity to process raw image-light pairs.

Furthermore, the feature map output by the deep extractor contains high-level semantic representations, which is crucial for improving the accuracy of normal reconstruction and preserving the details of the normal map in complex regions.
When updating on this stage's predictions, we implement comprehensive update by unfreezing all extractors, optimizing all extractors.

By adaptively controlling gradient flow across stages, this strategy significantly reduces error propagation while enhancing feature quality. It achieves more precise normal reconstruction than conventional approaches that uniformly update all network parameters throughout training.

\section{Experiments}
\subsubsection{Implementation Details}
To test the validity of our proposed approach, we conducted several experiments on three authoritative datasets. 
For quantitative evaluation, we use the mean angular error (MAE) in degrees to evaluate the performance of the estimated surface normals, as follows:
\begin{equation}
    MAE = \frac{1}{HW} \sum_{i,j}^{HW} (arccos( \tilde{N}_{i,j} \cdot N_{i,j}))
\end{equation}
where $\tilde{N}_{i,j}$and $N_{i,j}$represent the estimated normal and the true normal, respectively, and $HW$ represents the spatial resolution of the image.
At the same time, we also use $err^{15\degree}$ and $err^{30\degree}$ to 
indicate the percentage of all surface pixels with an angle error less than 15\degree and 30\degree,  respectively ($\%$) \cite{ERR15-ju_lam_chen_qi_dong_2020}.
Our model has only 2.2 million learnable parameters.
We apply Blobby \cite{Blobby-Johnson_Adelson_2011}, Sculpture dataset \cite{Sculpture-Wiles_Zisserman_2017} for training, and DiLiGenT
benchmark \cite{th2846_8281537}, DiLiGenT-$\mathrm{\Pi}$ \cite{diligent-pi_Wang_Ren_Guo_2023_ICCV}, DiLiGenT$10^{2}$ \cite{diligent100_Ren_2022_CVPR} for evaluating. 
The training process was conducted on a single NVIDIA GeForce RTX 3090 24GB GPU for around 12 hours.

% 
% ---修改后的版本---
% Our model has only 2.2 million learnable parameters. Training was conducted on a single NVIDIA GeForce RTX 3090 24GB GPU and took approximately 12 hours to complete. Specifically, we trained the model for 30 epochs using the Adam optimizer ($\beta_1$ = 0.9, $\beta_2$ = 0.999) with an initial learning rate of 0.00211, which was halved every 5 epochs.
% % 
% We apply Blobby \cite{Blobby-Johnson_Adelson_2011}, Sculpture dataset \cite{Sculpture-Wiles_Zisserman_2017} for training, and DiLiGenT
% benchmark \cite{th2846_8281537}, DiLiGenT-$\mathrm{\Pi}$ \cite{diligent-pi_Wang_Ren_Guo_2023_ICCV}, DiLiGenT$10^{2}$ \cite{diligent100_Ren_2022_CVPR} for evaluating. 
% % 

\begin{table}[]
\centering
\caption{Ablation results on DiLiGenT benchmark \cite{th2846_8281537}, in terms of MAE, the proportion of pixels with MAE less than $15^\circ$ and $30^\circ$. MFF: Multi-stage Feature Fusion; MFE: Multi-stage Feature Extraction; SUS: Selective Update Strategy.}
\label{ablation_table}
\renewcommand{\arraystretch}{1.2}
\setlength{\tabcolsep}{10pt}
\begin{tabular}{|c|c|c|c|c|c|c|}
\hline
\multicolumn{1}{|c|}{ID} & \multicolumn{1}{c|}{MFE} & \multicolumn{1}{c|}{MFF} & \multicolumn{1}{c|}{SUS} & \multicolumn{1}{c|}{MAE $\downarrow$} & \multicolumn{1}{c|}{$err^{15^\circ}$$\uparrow$} & \multicolumn{1}{c|}{$err^{30^\circ}$$\uparrow$} \\ 
\hline
(0) & -- & -- & -- & 6.78 & 91.9\% & 98.0\% \\
\hline
(1) & \checkmark & -- & -- & 6.60 & 91.8\% & 98.0\% \\
\hline
(2) & \checkmark & -- & \checkmark & 6.58 & 92.2\% & 98.1\% \\
\hline
(3) & -- & \checkmark & -- & 6.77 & 91.7\% & 98.0\% \\
\hline
(4) & \checkmark & \checkmark & -- & 6.66 & 91.9\% & 98.0\% \\
\hline
(5) & \checkmark & \checkmark & \checkmark & \textbf{6.23} & \textbf{92.3\%} & \textbf{98.1\%} \\
\hline
\end{tabular}
\end{table}

% 
% 消融实验
\subsection{Ablation study}
We systematically performed an ablation study of our model using the DiLiGenT benchmark \cite{th2846_8281537}.
% 直角双引号示例: ` `Hello, world!'' 
% 弯角双引号示例: ``Hello, world!'' 
Since the first 20 images of the ``Bear'' object are corrupted, only the last 76 images of this object were used \cite{CNN_PS_Ikehata_2018}.
All 96 images were used for the other nine objects.
Table \ref{ablation_table} shows the results of the different ablation experiments.
A lower MAE indicates better performance,while higher values for $err^{15\degree}$ and $err^{30\degree}$ are better.
% 
% 

% 啥也没有
% fusion 多阶段回归 锁住梯度
% 0 0 0
In ID(0), the multi-stage feature extraction and the selective update strategy used in conjunction with it are removed.
The multi-stage feature fusion is also removed and only the feature maps are connected by simple concatenation operations.
% 
% 0 1 0
ID(1) adds to ID(0) the multi-stage feature extraction.
% 加入多次法线回归、锁住梯度
% 0 1 1
ID(2) then adds to ID(1) the selective update strategy.
% 加入fusion模块
% 1 0 0
ID(3) adds the multi-stage feature fusion on the basis of ID(0).
% 加入多次法线回归、fusion
% 1 1 0
ID(4) is based on ID(3), adding the multi-stage feature extraction.
% 加入多次法线回归、锁住梯度、fusion
% 1 1 1
ID(5) is based on ID(4), adding the selective update strategy.
It is worth noting that the selective update strategy is based on multi-stage feature extraction, so it cannot be used without using multi-stage feature extraction.

\subsubsection{Effectiveness of Multi-stage Feature Extraction}
% Comparison of id(0) and id(1)
% Comparison of id(3) and id(4)

No matter the comparison results of ID(0) and ID(1), or ID(3) and ID(4), it can be seen that MAE has relatively obvious improvement.
The model can analyze the feature information of different orders by using multi-stage feature extraction. 
Because the available information is more sufficient, the quality of reconstructed normal diagrams has been improved significantly.

However, the improvement of $err^{15\degree}$and $err^{30\degree}$is not significant.
In other words, without the selective update strategy, the multi-stage feature extraction alone can not significantly improve the model.
\subsubsection{Effectiveness of Multi-stage Feature Fusion}
% ID(0)和ID(3)的对比
The comparison between ID(0) and ID(3) shows that the MAE performance of the model is slightly improved by adding the multi-stage feature fusion, but the other two indicators are basically unchanged.
% ID(1)和ID(4)的对比
The comparison results of ID(1) and ID(4) shows that without using our proposed selective update strategy, only using multi-stage feature extraction, the fusion does not significantly improve the model performance, and even slightly reduces the model performance.

% ID(2)和ID(5)的对比
A comparison of ID(2) and ID(5) shows that,
after incorporating selective update strategy, the multi-stage feature fusion enables the model to make full use of non-maximal features through the interaction between feature maps, and the MAE improvement is obvious.

\subsubsection{Effectiveness of Selective Update Strategy}
% 
% ID(1) vs ID(2): 多阶段特征+选择更新策略的影响
When comparing ID(1) and ID(2), multi-stage feature extraction coupled with our selective update strategy yields significant surface reconstruction improvements.

% ID(4) vs ID(5): 特征融合下的增量改进
The ID(4)-to-ID(5) comparison reveal that adding selective update strategy atop multi-stage feature extraction and multi-stage feature fusion substantially improves MAE, yet induces minimal gains in $err^{15\degree}$ and $err^{30\degree}$. 
This indicates that accuracy improvements primarily occur in regions with errors greater than $30\degree$, underscoring persistent challenges in high-error areas.

\begin{figure}[!ht]  
    \centering
    \subfloat[ ]{
        \includegraphics[width=0.5\textwidth]{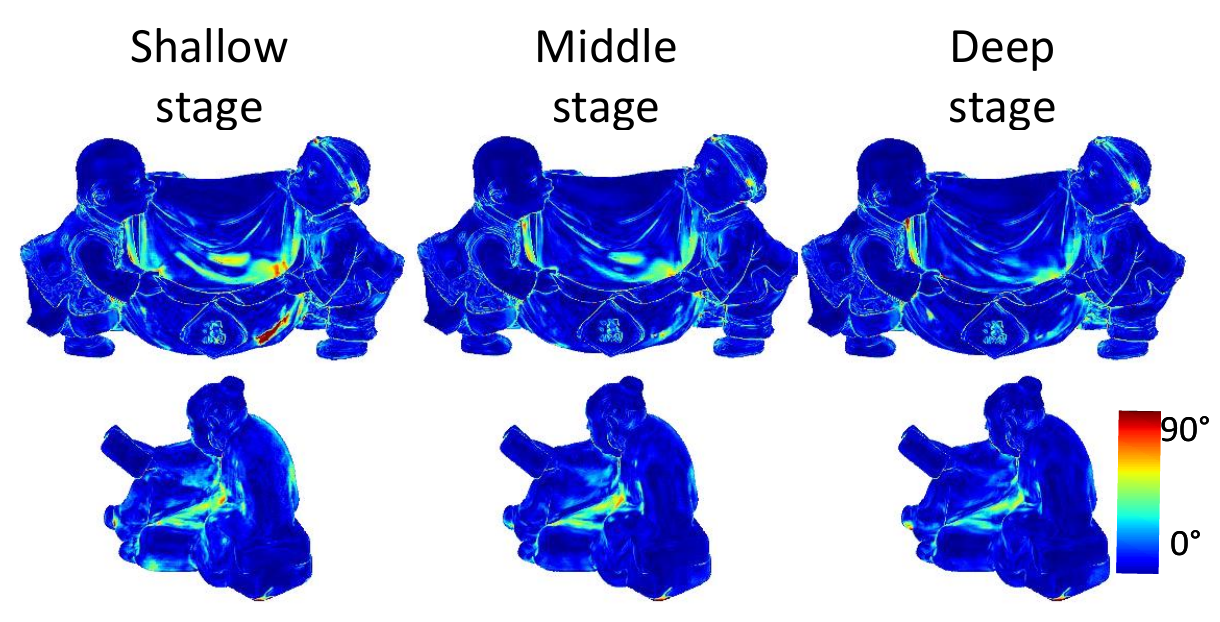}
        \label{multi_stage_results}
    }
    \hfill
    \subfloat[ ]{
        \includegraphics[width=0.4\textwidth]{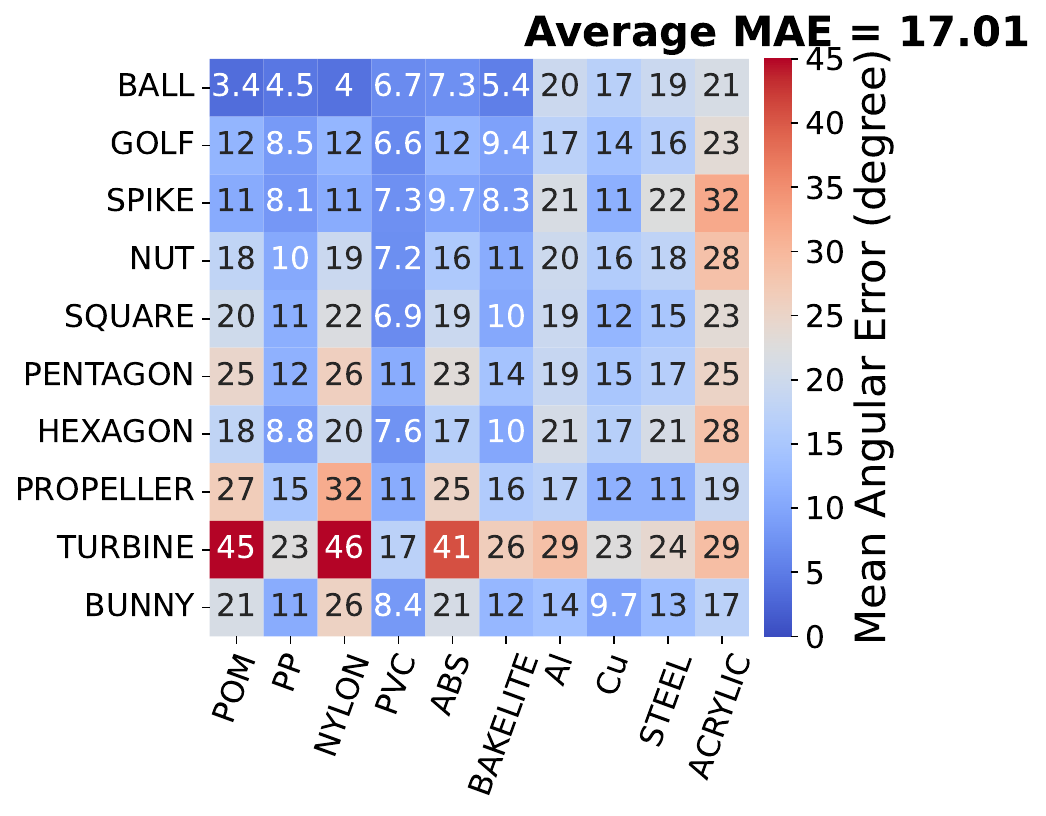}
        \label{diligent100_result}
    }
    
    \caption{
    (a) is the qualitative results at different stages shows that the quality of the predicted normals increases with the stage levels.
    (b) is the result of our model on DiLiGenT$10^{2}$ Dataset, A number in each element of the matrix indicates an MAE in degrees.
        }
    \label{dataset-result}
\end{figure}

% 多阶段法线可视化证据
As validated by DiLiGenT dataset \cite{th2846_8281537} results in Fig. \ref{multi_stage_results}, selective update strategy drives progressive refinement of normal maps across stages.

% 核心技术洞见
Thus, under multi-stage feature fusion, the selective update strategy enhances the model's capability to handle geometrically complex regions rather than merely refining simple-structure reconstruction. 
This directly strengthens the model's generalization capacity.
\subsubsection{Weight Selection of the Loss Function at different stages}
Multi-stage normal predictions exhibit inherent quality progression: shallow features capture original characteristics but lack precision, while deep features synthesize comprehensive surface details through hierarchical processing. 
This structural reality necessitates loss weighting, which strengthens the model's representational capacity.

As shown in table \ref{loss_weight}, 0.5:0.5:1 weighting achieves significant gains in MAE, $err^{15\degree}$ and $err^{30\degree}$, evidencing reduced reconstruction uncertainty— a particularly valuable in non-Lambertian regions where coarse predictions often underperform.

% 实验结果图（多阶段法线图和100数据集结果）

% \begin{figure}[!ht]  
%     \begin{subfigure}[b]{0.6\textwidth}
%         \includegraphics[width=\linewidth]{original_materials/figures/multi_stage_errorMapV3.pdf}
%         \caption{}
%         \label{multi_stage_results}
%     \end{subfigure}
%     % \hfill  % 水平填充，使两图均匀分布
%     \begin{subfigure}[b]{0.4\textwidth}
%         \includegraphics[width=\linewidth]{original_materials/figures/dataset100_result.pdf}
%         \caption{}
%         \label{diligent100_result}
%     \end{subfigure}
    
%     \caption{
%     (a) is the qualitative results at different stages shows that the quality of the predicted normals increases with the stage levels.
%     (b) is the result of our model on DiLiGenT$10^{2}$ Dataset, A number in each element of the matrix indicates an MAE in degrees.
%         }
%     \label{dataset-result}
% \end{figure}

% 
% 权重选择结果
\begin{table}[]
\renewcommand{\arraystretch}{1.27}
\setlength\tabcolsep{2.7pt}   %
\caption{Weight selection of the loss function at different stages.The best result is indicated in bold.} % 添加表注
\centering % 让表格居中
\footnotesize % 改变接下来文本的大小
% Note: loss weighted results 

\begin{tabular}{|c|c|c|c|c|c|}% 开始tabular环境，并指定三列都居中对齐
\hline
%  \makecell 命令，可以在单元格中插入多行文本。
\makecell{ weight of shallow} & 
\makecell{ weight of middle} & 
\makecell{ weight of deep} &
MAE & 
$err^{15\degree}$$\uparrow$ & $err^{30\degree}$$\uparrow$ \\ 
\hline
0.3&0.5&1& 6.58 &92.1\% &\textbf{98.1\%} \\
\hline
0.5&0.9&1& 6.65 &91.8\% &98.0\%\\
\hline
0.75&0.75&1& 6.67 & 91.2\%&97.9\%\\
\hline
0.5&0.5&1& \textbf{6.23} & \textbf{92.3\%}&\textbf{98.1\%}\\
\hline
\end{tabular}
\label{loss_weight} % 添加标签，以便在文档中引用此表
% 
% \footnotesize % 改变接下来文本的大小
% Note: This is additional information about the table. % 表注的附加信息
\end{table}

\subsection{DiLiGenT Benchmark Comparisons}
We evaluate our method against several state-of-the-art approaches on the DiLiGenT benchmark \cite{th2846_8281537}. 
As shown in Table \ref{diligent_result}, our model achieves the best or second-best performance on 4 out of the 10 objects, with an average Mean Angular Error (MAE) of 6.23 across all objects. 
This result not only demonstrates the superiority of our method over existing SOTA approaches but also highlights its ability to generalize across different object geometries and material properties.

\begin{figure}[!t]
    \centering
    \includegraphics[width=\textwidth]{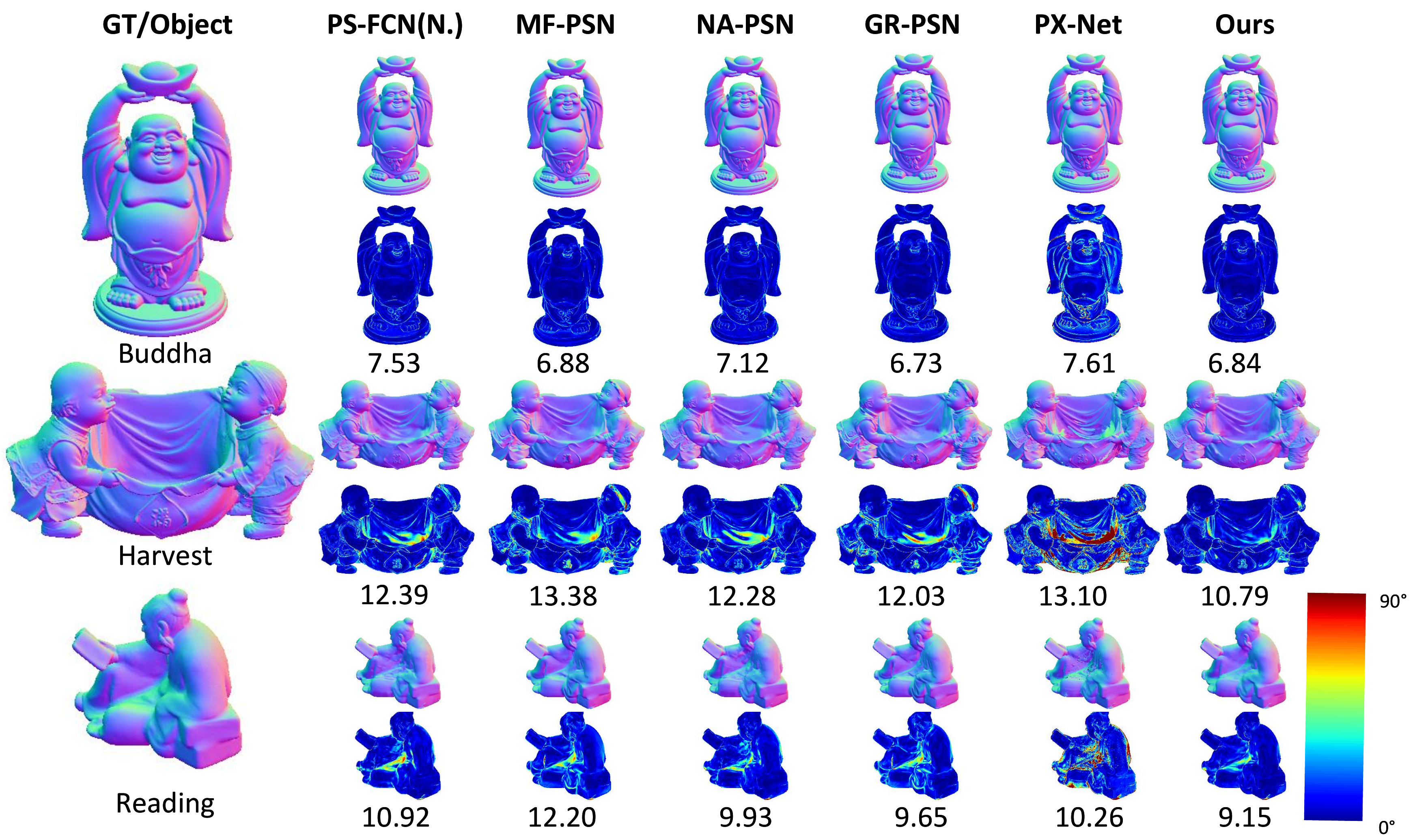}
    \caption{
    Qualitative results for objects ``Buddha", ``Harvest" and ``Reading" on the DiLiGenT benchmark dataset with 96 input images.
    The number below the error map is the MAE of degrees. 
    By comparing with
    PS-FCN(N.) \cite{psfcn-n_Chen2020DeepPS}, 
    MF-PSN \cite{mf_psn-Liu_Ju_Jian_Gao_Rao_Hu_Dong_2022}, 
    NA-PSN \cite{NA_PSN-Ju_Shi_Jian_Qi_Dong_Lam_Kenneth}, 
    GR-PSN \cite{GR_PSN-10306333}, 
    and PX-Net \cite{PX_Net-Logothetis_Budvytis_Mecca_Cipolla_2020}, 
    our model achieves the best or second-best performance. 
    }
    \label{errormap_visible}
\end{figure}

Visual comparisons on challenging objects (``Buddha", ``Harvest", ``Reading") in Fig. \ref{errormap_visible} further validate our advantages. 
Specifically, our method significantly improves normal reconstruction in critical regions (e.g., Buddha's shoulder, Harvest's pocket, Reading's clothing). 
This enhancement stems from our selective update strategy, which boosts feature extraction in complex areas with shadows, highlights, and inter-reflections.

Collectively, our approach delivers competitive quantitative results while excelling in fine-detail reconstruction for challenging photometric stereo scenarios.

% DiLiGenT结果对比
\begin{table}
\centering % 让表格居中

\footnotesize % 改变接下来文本的大小
\caption{
Comparisons on DiLiGenT-main benchmark \cite{th2846_8281537}.
Boldface indicates the best result and underline indicates the second best result.} % 添加表注
\resizebox{\textwidth}{!}{
\begin{tabular}{|c|cccccccccc|c|}
\hline
Method & Ball & Bear & Buddha & Cat & Cow & Goblet & Harvest & Pot1 & Pot2 & Reading & Avg. \\
\hline
% L2 \cite{Woodham_1980} & 4.10 & 8.39 & 14.92 & 8.00 & 25.60 & 18.50 & 30.62 & 8.89 & 14.65 & 19.80 & 15.39 \\
% PS-FCN(N.) \cite{psfcn-n_Chen2020DeepPS} & 2.67 & 7.72 & 7.53 & 4.76 & 6.72 & 7.70 & 12.39 & 6.17 & 7.15 & 10.92 & 7.39\\
MF-PSN \cite{mf_psn-Liu_Ju_Jian_Gao_Rao_Hu_Dong_2022} & 2.07 & 5.83 & 6.88 & 5.00 & 5.90 & 7.46 & 13.38 & 7.20 & 6.81 & 12.20 & 7.27\\
CNN-PS \cite{CNN_PS_Ikehata_2018} & 2.12 & 4.38 & 8.07 & 4.38& 7.92 & 7.42 & 14.08 & 5.37 & 6.38 & 12.12 & 7.20\\
PSMF-PSN \cite{PSMF_PSN-10301617} & 2.54 & 5.99 & 7.21 & 5.09 & 5.52 & 7.75 & 11.40 & 6.91 & 6.11 & 10.01 & 6.85\\
CFR-PSN \cite{coarse-to-fine_10688188} & 2.08 & 5.65 & 7.28 & \textbf{4.19} & 7.93 & 7.34 & 11.66 & \underline{5.27} & 6.70 & 9.98 & 6.81\\
SR-PSN \cite{sr_psn-10208243} & 2.23 & 5.24 & \underline{6.75} & 4.63 & 6.12 & 7.07 & 12.61 & 5.88 & 6.44 & 10.35 & 6.73\\
NA-PSN \cite{NA_PSN-Ju_Shi_Jian_Qi_Dong_Lam_Kenneth} & 2.93 & 4.65 & 7.12 & 4.65 & 5.99 & 7.49 & 12.28 & 5.96 & 6.42 & 9.93 & 6.72\\
GR-PSN \cite{GR_PSN-10306333} & 2.22 & 5.61 & \textbf{6.73} & 4.33 & 6.17 & \underline{6.78} & 12.03 & 5.54 & 6.42 & 9.65 & 6.55\\
IGA-PSN \cite{IGA-PSN_10.1007/978-981-96-0122-6_25} & 2.20 & 4.83 & 7.08 & 4.63 & 5.92 & 7.25 & \underline{11.01} & 5.85 & 6.41 & 9.37 & 6.46\\
PX-Net \cite{PX_Net-Logothetis_Budvytis_Mecca_Cipolla_2020} & \underline{2.03} & \underline{4.13} & 7.61 & 4.39 & \underline{4.69} & 6.90 & 13.10 & \textbf{5.08} & \underline{5.10} & 10.26 & 6.33\\
PS-SINI \cite{PS-SINI_Yang2024NonLambertianPS} & \textbf{1.90} & \textbf{3.61} & 7.69 & 4.43 & \textbf{4.52} & \textbf{6.63} & 15.23 & 5.78 & \textbf{4.85} & \textbf{8.27} & \underline{6.29}\\
MSF-Net(Ours) & 2.27 & 4.60 & 6.84 & \underline{4.20} & 6.08 & 7.40 & \textbf{10.80} & \textbf{5.08} & 5.85 & \underline{9.15} & \textbf{6.23} \\
\hline
\end{tabular}
}
\label{diligent_result} % 添加标签，以便在文档中引用此表
\end{table}

% 

% 是大写的Π
\subsection{DiLiGenT-$\mathrm{\Pi}$ Dataset Comparisons}
Accurate reconstruction of shape details is a critical challenge in photometric stereo, particularly for objects with intricate features commonly found in real-world scenes, such as reliefs, badges, and coins. To rigorously evaluate the performance of our model in close-range surface reconstruction, we conducted a comprehensive comparative analysis against several state-of-the-art methods using the DiLiGenT-$\mathrm{\Pi}$ dataset \cite{diligent-pi_Wang_Ren_Guo_2023_ICCV}.

The evaluation encompasses Least squares (L2 \cite{Woodham_1980}), three per-pixel strategies (CNN-PS \cite{CNN_PS_Ikehata_2018}, GPS-Net \cite{GPS_Net-Yao_Li_Fu_Hu_Shi_2020}, PX-Net \cite{PX_Net-Logothetis_Budvytis_Mecca_Cipolla_2020}), and two full-pixel methods (PS-FCN \cite{PSFCN_Chen_2018}, NA-PSN \cite{NA_PSN-Ju_Shi_Jian_Qi_Dong_Lam_Kenneth}). The dataset is categorized into four material reflectance groups—metal, mirror, rough, and translucent—providing a diverse benchmark for evaluation.

As shown in Table \ref{pi_result}, our model achieves best or second-best on nearly half of objects, excelling particularly on mirror and rough surfaces. 
This success can be attributed to the model's ability to effectively capture fine-grained details and complex geometries under varying lighting conditions. 
Notably, our model achieves these competitive results with a significantly smaller parameter count compared to many state-of-the-art methods, highlighting its efficiency and suitability for resource-constrained applications.

% diligent误差图对比

\begin{table}[!bt]
% \scriptsize
% \renewcommand{\arraystretch}{1} % 变大行间距
% \setlength\tabcolsep{1pt}%调列距
\renewcommand{\arraystretch}{1}
\setlength\tabcolsep{0.3pt}%调列距
    \centering
    \caption{benchmark results on our real-world dataset DiLiGenT-$\mathrm{\Pi}$ \cite{diligent-pi_Wang_Ren_Guo_2023_ICCV}. We calculate the average MAE of the object in each material.}
    \begin{tabular}{|c|cccccc|c|}
    \hline
    material & L2 \cite{Woodham_1980} & 
    PX-Net \cite{PX_Net-Logothetis_Budvytis_Mecca_Cipolla_2020} & 
    PS-FCN  \cite{PSFCN_Chen_2018} & 
    GPS-Net \cite{GPS_Net-Yao_Li_Fu_Hu_Shi_2020} & 
    CNN-PS \cite{CNN_PS_Ikehata_2018} & 
    NA-PSN \cite{NA_PSN-Ju_Shi_Jian_Qi_Dong_Lam_Kenneth} & 
    Ours \\
    \hline
    Metallic & 7.2 & 9.2 & 5.2 & \underline{5.0} & \textbf{4.9} & \textbf{4.9} & 5.4\\
    Specular & 8.5 & 8.4 & 8.5 & 8.9 & \textbf{7.4} & \underline{7.6} & 7.8\\
    Translucent & 17.0 & \textbf{16.0} & 17.3 & 17.6 & \underline{16.8} & 17.0 & 17.0\\
    Rough & 14.6 & 13.7 & 14.5 & 13.8 & \underline{13.5} & \textbf{13.3} & \textbf{13.3}\\
    \hline
    AVG. & 10.5 & 10.8 & 9.8 & 10.1 & \textbf{9.2} & \textbf{9.2} & \underline{9.4} \\
\hline
\end{tabular}
% \end{threeparttable}
\label{pi_result}
\end{table}

\subsection{DiLiGenT$10^{2}$ Dataset Comparisons}
For an in-depth analysis of our model's ability to generalize to different objects and materials, we evaluated MSF-Net on the challenging DiLiGenT$10^{2}$ dataset, as illustrated in Fig. \ref{diligent100_result}.
Our method achieved an average MAE of 17.01, outperforming most methods reported in their paper \cite{diligent100_Ren_2022_CVPR}.
These encompass traditional methods L2 \cite{Woodham_1980} (18.13), TH28 \cite{th2846_8281537} (19.66), TH46 \cite{th2846_8281537} (18.82), WG10 \cite{wg10_10.1007/978-3-642-19318-7_55}, ST14 \cite{BRDF_1-fShi_Tan_Matsushita_Ikeuchi_2014} (18.34), PF14 \cite{pf14_6247754} (30.63), and learning-based methods IRPS \cite{IRPS-pmlr-v80-taniai18a} (17.10), GPS-Net \cite{GPS_Net-Yao_Li_Fu_Hu_Shi_2020} (19.98).

% 解释下不足
However,  performance degradation was observed on POM and NYLON materials, primarily due to two factors.
First, the scarcity of these materials in the training dataset reducing the model's ability to generalize effectively to such surfaces. 
Second, while parameter efficiency improves inference speed, it restricts the model's capacity to capture complex surface characteristics of these materials.

Notably, despite these limitations, our model achieves competitive accuracy against state-of-the-art methods while utilizing significantly fewer parameters.

\section{Conclusions}
This paper presents a novel photometric stereo network that achieves high-quality normal reconstruction through multi-stage feature extraction and fusion. 
The core innovation lies in a selective update strategy, which progressively calibrates the feature extractor step by step, thus ensuring the ability to capture accurate information.
Our comprehensive ablation study confirmed the effectiveness of the proposed approach. 
By comparing the results of the DiLiGenT benchmark \cite{th2846_8281537}, the DiLiGenT-$\mathrm{\Pi}$ dataset \cite{diligent-pi_Wang_Ren_Guo_2023_ICCV} and the DiLiGenT$10^{2}$ dataset \cite{diligent100_Ren_2022_CVPR}, we successfully validate the performance of the method.
Notably, our model attains competitive accuracy with only 2.2M parameters, highlighting its efficiency for practical deployment.
Our code and models can be found at https://github.com/RainLikeCodes/MSF-Net.
Current limitations involve reconstructing normals for materials with complex optical properties (e.g., translucency and high specularity). 
Future work will focus on material-adaptive mechanisms to enhance generalization across diverse real-world conditions.

\bibliographystyle{splncs04}
% \bibliography{refs}
\bibliography{original_materials/main}

\end{document}